\pdfoutput=1

\documentclass[11pt]{article}

\usepackage[]{ACL2023}
\usepackage{times}
\usepackage{latexsym}
\usepackage{enumitem}
\usepackage{booktabs}
\usepackage{multirow}

\usepackage[T1]{fontenc}

\usepackage[utf8]{inputenc}

\usepackage{microtype}

\usepackage{inconsolata}

\usepackage{graphicx}
\usepackage{subcaption}
\usepackage{amsmath}

\usepackage[compact]{titlesec}
\usepackage{hyperref}
\newcommand{\vh}{\mathbf{h}}

\newcommand{\method}{\textsc{Diffusion Lens}}

\title{Diffusion Lens: Interpreting Text Encoders in Text-to-Image Pipelines}

\author{Michael Toker\hspace{1em} Hadas Orgad \hspace{1em} Mor Ventura \hspace{1em}  {\bf Dana Arad} \hspace{1em}  {\bf Yonatan Belinkov} \\
 Technion -- Israel Institute of Technology \hspace{2em} 
  \\
        \texttt{\{tok,orgad.hadas,mor.ventura,danaarad,belinkov\}@campus.technion.ac.il} \\
        }

\begin{document}

\setlength{\abovedisplayskip}{3pt}
\setlength{\belowdisplayskip}{4pt}

\maketitle

\begin{abstract}

Text-to-image diffusion models (T2I) use a latent representation of a text prompt to guide the image generation process.
However, the process by which the encoder produces the text representation is unknown.
We propose the \method, a method for analyzing the text encoder of T2I models by generating images from its intermediate representations.
Using the \method, we perform an extensive analysis of two recent T2I models.
Exploring compound prompts, %
we find that complex scenes describing multiple objects are composed progressively and more slowly compared to simple scenes; 
Exploring knowledge retrieval, we find that representation of uncommon concepts require further computation compared to common concepts,
and that knowledge retrieval is gradual across layers.
Overall, our findings provide valuable insights into the text encoder component in T2I pipelines.\footnote{Code and data are available on the project webpage \href{https://tokeron.github.io/DiffusionLensWeb/}
{{tokeron.github.io/DiffusionLensWeb}}.}

\end{abstract}

\section{Introduction}
\label{sec:intro}

\begin{figure}[ht]
\centering
\includegraphics[width=\linewidth]{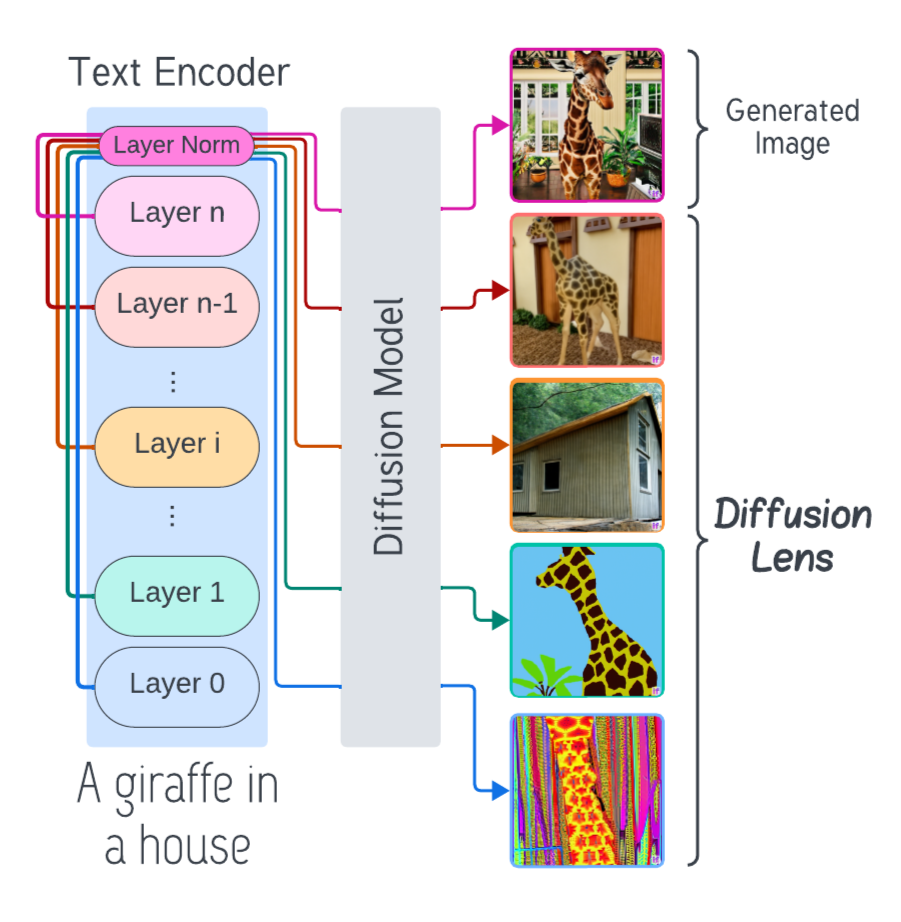}
\caption{Visualization of the text encoder's intermediate representations using the \method. At each layer of the text encoder (in blue), the \method~takes the full hidden state, passes it through the final layer norm, and feeds it into the diffusion model.}
\label{fig:intro_fig}
\end{figure}

The text-to-image (T2I) diffusion pipeline comprises two main elements: the text encoder and the diffusion model.
The text encoder converts a textual prompt into a latent representation, while the diffusion model utilizes this representation to generate the corresponding image.
Several recent studies have delved into the internal workings of the diffusion model and the cross-attention mechanism that connects the two components \cite{tang2022daam, hertz2023prompttoprompt, orgad2023editing, chefer2023attend}.
Yet,  while the text encoder is a key component of the pipeline with a large effect on image quality and text-image alignment \cite{saharia2022photorealistic}, its internal mechanisms remain unexplored.
Moreover, while there is a wide range of work that has analyzed general language model internals \cite{belinkov2019analysis,Rogers2020API,madsen2022post}, these methods are not suitable for exploring fine-grained visual features.

We propose the \method, a method for analyzing the inner mechanism of the text encoder.
The \method~uses intermediate representations of the prompt from various layers of the text encoder to guide the diffusion process, resulting in images that are clear, consistent, and easy to understand for most layers (see Figure \ref{fig:intro_fig}).
Notably, the \method~relies solely on the pre-trained weights of the model and does not depend on any external modules.

\begin{figure*}[t]
 \centering
\includegraphics[width=\textwidth]{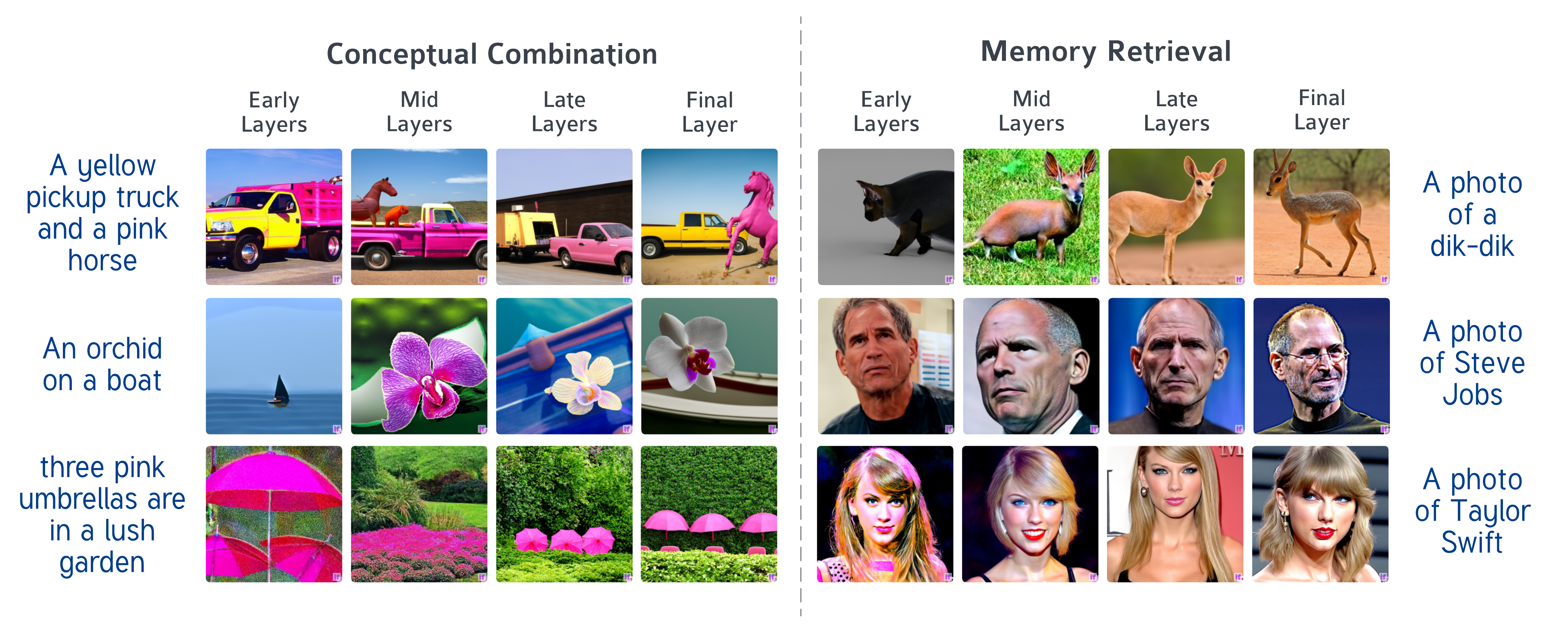}
 \caption{Insights gained from using \method. \textbf{Conceptual Combination} (left): early layers often act as a ``bag of concepts'', lacking relational information which emerges in later layers. \textbf{Memory Retrieval} (right): uncommon concepts gradually evolve over layers, taking longer to generate compared to common concepts.}
 \label{fig:examples}
\end{figure*}

We employ the \method~to examine the computational process of the text encoder in two popular T2I models: Stable Diffusion \cite{rombach2022high} and Deep Floyd \cite{deepfloyd}.
Our investigation focuses on two main analyses: the model's capability of conceptual combination and its memory retrieval process.
For each analysis, we either construct a tailored dataset to isolate a specific phenomenon or utilize naturally occurring human-written image captions.

Our analysis of conceptual combination reveals various insights:
(1) Complex prompts such as ``A yellow pickup truck and a pink horse'' require more computation to achieve a faithful representation compared to simpler prompts such as ``A green cat''.
(2) Complex representations are built gradually: as illustrated in Figure \ref{fig:examples} (Left), images generated from early layer representations typically encode concepts separately or together without capturing their correct relationship, resembling more of a ``bag of concepts''. Images from subsequent layers encode the relationships more accurately.
(3) The order in which objects emerge during computation is influenced by either their linear or syntactic precedence in the sentence. Here, we find a difference between the two examined models: Deep Floyd's text encoder, T5 \cite{raffel2020exploring}, shows a greater sensitivity to syntactic structure, while Stable Diffusion's text encoder, CLIP \cite{clip2021}, tends to reflect linear order.

Next, we investigate memory retrieval, and uncover several key findings:
(1) Common concepts, such as ``Kangaroo'', emerge in early layers while less common ones, such as the animal ``Dik-dik'', gradually emerge across the layers, with the most accurate representations predominantly occurring in the upper layers, as illustrated in Figure \ref{fig:examples} (Right, top).
(2) Fine details, like human facial features, materialize at later layers, as shown in Figure \ref{fig:examples}, with the prompt ``A photo of Steve Jobs''.
(3) Knowledge retrieval is gradual, unfolding as computation progresses. This observation diverges from prior research on knowledge encoding in language models which characterizes knowledge as a localized attribute encoded in specific layers \cite{geva-etal-2022-transformer, meng2022locating, arad2023refact}. 
(4) Notably, there are discernible differences in memory retrieval patterns between the two text encoders: Deep Floyd's T5 memory retrieval exhibits a more incremental behavior compared to Stable Diffusion's CLIP.
The disparities uncovered through our analyses suggest that factors such as architecture, pretraining objectives, or data may influence the encoding of knowledge or language representation within the models.

Our contributions are summarized as follows: 
\begin{itemize}[itemsep=1pt,topsep=1pt,parsep=1pt]
\item We develop the \method, a new intrinsic method for analyzing the intermediate states of the text encoder within T2I pipelines. 
\item We conduct thorough experiments that reveal insights on the computational mechanisms of text encoders in the T2I pipeline. Our findings shed light on how factors such as complexity, frequency, and syntactic structure impact the encoding process.
\end{itemize}

\section{Diffusion Lens}
\label{sec:diffusion_lens}
\paragraph{Preliminiaries.}
Current text-to-image diffusion models comprise two main components \cite{ saharia2022photorealistic, ramesh2022hierarchical}: a language model used as a text encoder that takes the textual prompt as input and produces latent representations; and a diffusion model that is conditioned on the representations from the text encoder and generates an image from an initial input noise.

The language model in the T2I pipeline is typically a transformer model. Transformer models consist of a chain of transformer blocks, each composed of three sub-blocks: attention, multi-layer perceptron, and layer norm \cite{vaswani2017attention}. 

We denote the transformer block at layer $l$ as $F_l$. The input to the model is a sequence of $T$ word embeddings, denoted as $\smash{\vh^0 = [h^0_1,\ldots,h^0_T]}$. Then, the output of the transformer block at layer $l$ is a sequence of hidden states $\vh^{l+1}$:
\begin{equation}
 \vh^{l+1} = F_l(\vh^l)
\end{equation}

The output representations of the last block, $L$, go through a final layer norm, denoted as $ln_f$. Then, they condition the image generation process through cross-attention layers, resulting in an image $I$. We abstract this process as: 
\begin{equation}
 I = \text{Diff}(ln_f(\vh^L))
\end{equation}

\paragraph{Diffusion Lens.}

In a T2I pipeline with a text encoder of $L$ layers, for layer $l < L$, we process the output of block $l$, including padding tokens, through the final layer norm. We condition the diffusion process on this output, as illustrated in Figure \ref{fig:intro_fig}.
Namely, we generate an image $I$ from an intermediate layer $l$ as follows: 

\begin{equation}
I = \text{Diff}(ln_{f}(\vh^{l}))
\end{equation}
The final layer norm is a crucial step in generating coherent images (see Appendix \ref{app:final_layer_norm}).
It projects the representations into the cross-attention embedding space without the caveat of adding new information to the representation, as may happen with learned projections. This process generates an image representing the intermediate state of the text-encoder as interpreted by the diffusion model.

\section{Experimental Setup}
\label{sec:experimental_setup}

\paragraph{Models.}
The experiments are performed on Stable Diffusion 2.1 \cite[denoted \emph{SD},][]{rombach2022high} and Deep Floyd \cite[denoted \emph{DF},][]{deepfloyd}. 
SD is an open-source implementation of latent diffusion \cite{rombach2022high}, with OpenCLIP-ViT/H \cite{ilharco_gabriel_2021_5143773} as the text-encoder.
DF is another open-source implementation of latent diffusion inspired by \citet{saharia2022photorealistic}, with a frozen T5-XXL \cite{raffel2020exploring} as the text encoder.
We usually only report the results on DF, unless there is a difference between the models, which we then discuss. The full results on SD are given in Appendix \ref{app:sd_results}.

\paragraph{Data.}
Depending on the specific experiment, we either curate prompt templates and automatically generate a list of prompts from a collected list of concepts we are interested in investigating, or use a list of natural, handwritten prompts from COCO \cite{lin2015microsoft}. The data for each experiment is detailed in the next sections.
With each prompt, we generate images that are conditioned on representations from every fourth layer in the model, which serves as a representative subset. This results in 7 images for DF (which has 25 layers in total) and 6 images for SD (which has 24). We generate each prompt using four seeds. 

\paragraph{Evaluation.}
In every experiment we ask questions about the images at every layer, e.g., ``Does the prompt correspond to the generated image''; or, if there are two objects in the prompt, ``Does object A appear in the generated image?''. We describe the questions in detail for every experiment below.
To analyze the representation building process of successful generations, we report our main findings on cases where all the images from the last layer align with the prompt. We separately analyze model failures in Section \ref{sec:error_analysis}.

We annotated the generated images using both human annotators and \mbox{GPT-4V} \cite{openai2023gpt4}. For the human evaluation, we collected answers to the questions by ten human annotators, with 10\% overlap to measure inter-annotator agreement.
We found a high agreement between \mbox{GPT-4V} and the humans (Table~\ref{tab:annotations_agreement}, App.~\ref{app:annotation}).
We provide the main results based on the human annotations; however, our results support the use of automatic annotation to allow larger scale and reduced cost. 
Overall, we collected answers to roughly $66,560$ questions, 37\% of them by GPT-4V. 
For full details on the annotation process, inter-annotator agreement, and integration with GPT-4V, refer to Appendix \ref{app:annotation}.

\section{Conceptual Combination}

T2I diffusion models are popular for their ability to generalize beyond their training data, creating composite concepts \cite{ramesh2022hierarchical}.
Conceptual combination is the cognitive process by which at least two existing basic concepts are combined to generate a new higher-order, composite concept \cite{WU2009173}.
Conceptual combination is at the core of knowledge representation, since it asks how the meaning of a complex phrase connects to its component parts \cite{hampton2013conceptual}, e.g., ``A cat in a box''.
This section uses the \method~to trace the process by which the text encoder creates composite concepts.

\begin{figure}[t]
 \centering
 \includegraphics[width=\linewidth]{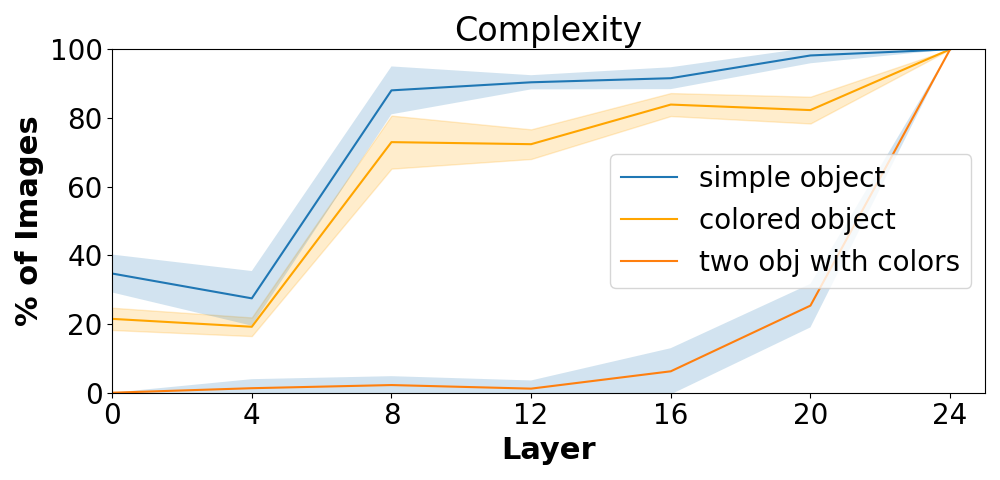}
 \caption{Percentages of prompt-matching images across various layers. As prompts become more complex, \method~ has to utilize more layers to extract a correct image.}
 \label{fig:complex_comes_later}
\end{figure}

\subsection{Building complex scenarios}

This study investigates the text encoder's ability to combine concepts at varying levels of complexity. We utilize COCO classes \cite{lin2015microsoft} as a diverse set of prompts with readily identifiable visual meanings. Each experiment commences with a simple list of objects as prompts, progressively increasing in complexity as outlined subsequently.

\paragraph{Colors and conjunction.} We compile three lists of prompts:
(1) objects (e.g., ``a dog''); (2) objects with color description (``a red dog''); and (3) two objects with colors  (``a red dog and a white cat'').
To investigate how conceptual combination emerges through the layers,
we annotated a random sample of 80 prompts,\footnote{In this experiment, human annotators annotated 40 prompts and GPT4-V annotated an additional 40.} asking the following questions for each layer: (a) Does object X appear in the image? (b) Does color X appear in the image? (c) Does object X appear in the correct color?
X is either the 1st or the 2nd object, for a total of 6 questions.

\paragraph{Physical relations.} We compile two lists of prompts: (1) objects, (2) prompts describing two objects and a preposition: either ``in'' or ``on''. For example, ``A cat in a box''.
We sample 40 prompts. We ask three questions: (a-b) Does object X appear in the image? and (c) Is object A in/on object B?

\subsubsection*{Results}

\begin{figure}[t]
 \centering
 \includegraphics[width=\linewidth]{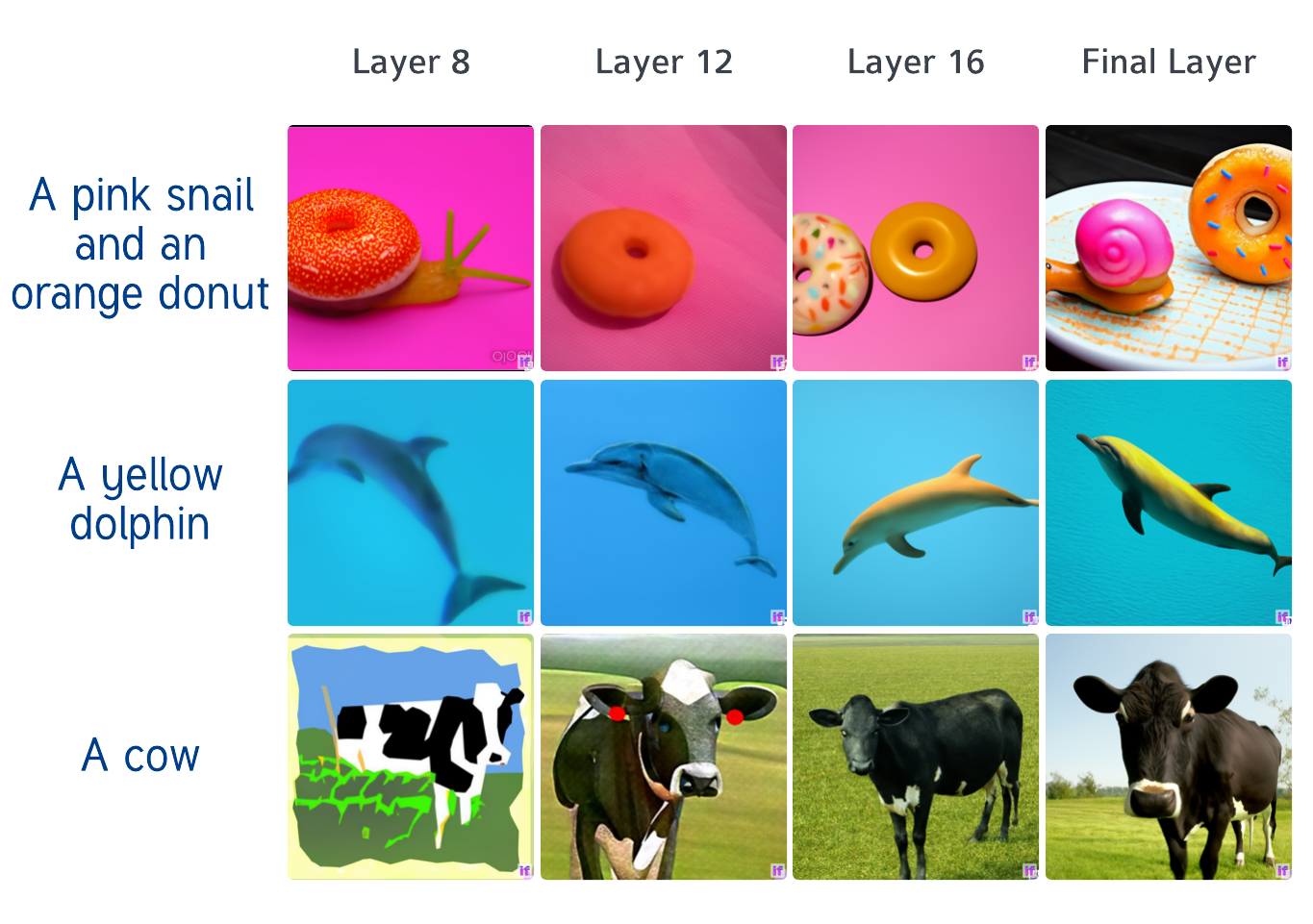}
 \caption{Complex prompts take more computation blocks to emerge.}
 \label{fig:complex_vs_simple}
\end{figure}

\paragraph{The simpler the concept, the earlier it emerges.} 
Figure \ref{fig:complex_comes_later} shows the percentage of images that correctly generated the concepts for each category: an object, an object and a color, and two colored objects. 
Prompts describing a single object emerge the earliest, between layers 4 and 16, while prompts containing a color descriptors emerge in layers 16--20. Conjunction prompts emerge last, around layers 20--24.
As demonstrated in Figure \ref{fig:complex_vs_simple}, ``A cow'' is fully represented by layer 8, while ``A yellow dolphin'' does not correctly form until layer 16. Lastly, ``A pink snail and an orange donut'' only fully forms at much later layers, correctly matching the objects and colors at the final layer, 24.

\begin{figure}[t]
 \centering
 \includegraphics[width=\linewidth]{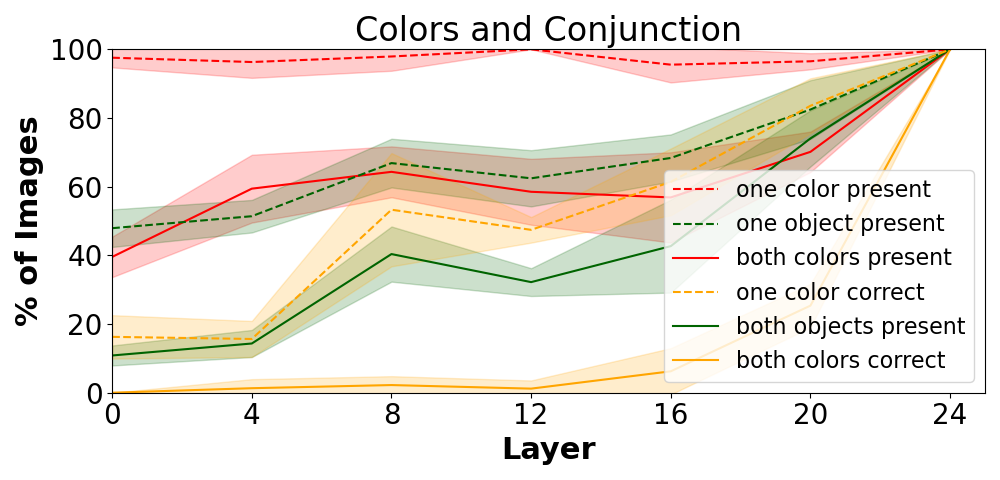}
 \caption{The proportion of images where either the object, the colors, or both were present, and where either the objects or the colors were accurately represented.}
 \label{fig:gradual_complexity}
\end{figure}

\begin{figure}
 \centering
 \includegraphics[width=\linewidth]{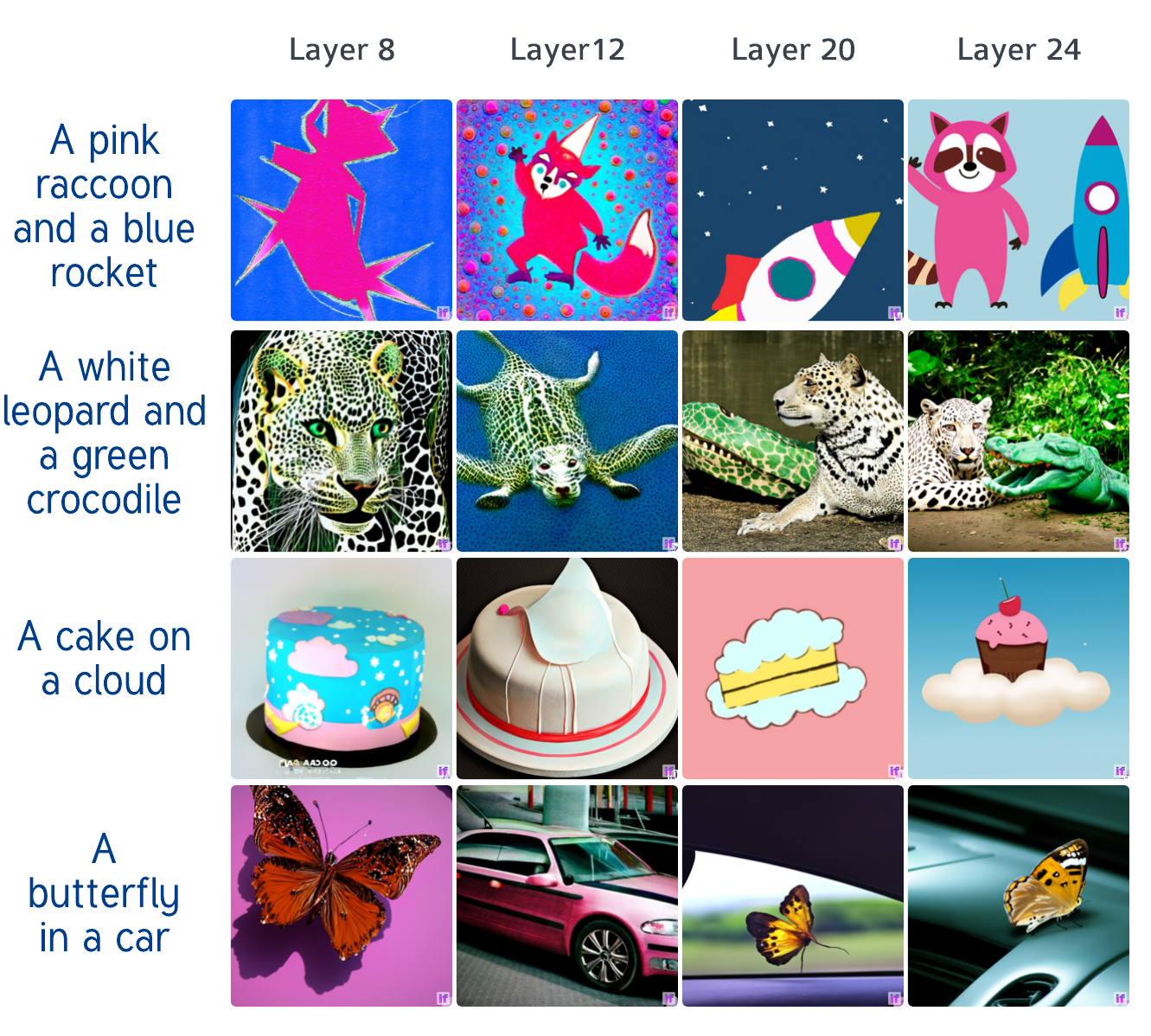}
 \caption{Complex representations are constructed gradually. In some cases, objects are mixed in early representations. In other cases, only one of the objects appear in early representations.}
 \label{fig:complex_gradally}
\end{figure}

 \paragraph{Complex representations are constructed gradually.}
 We continue with the complex prompts of two colored objects.
 Figure \ref{fig:gradual_complexity} aggregates the answers to illustrate the behavior of either or both objects appearing in intermediate layers. Colors often emerge first, with both colors often emerging in early layers in DF (in SD, the two objects appear before two colors). A single object is also gradually represented in layers 4-12. 
Notably, while the colors and one of the objects appear, the object is not necessarily generated in the correct color. This can be seen in the first example in Figure \ref{fig:complex_gradally}: While a raccoon and a rocket do appear, and the image contains both blue and pink elements, the rocket is not blue until the final layer.
In some cases, we observe a mixture of concepts in early layers, as seen in the second example of Figure \ref{fig:complex_gradally}.
Similarly, the bottom two examples in Figure \ref{fig:complex_gradally} show prompts composing two objects and a proposition. As with colors, we observe that individual objects appear in early layers but the correct relation emerges much later. For example, ``A cake on a cloud'' generates images of both a cake and a cloud, with different relations; at layer 8 the cake is decorated with a cloud and in layer 20 the clouds are depicted as frosting. The correct relation is only generated at the final layer. We provide a more detailed discussion of prepositions in Appendix \ref{app:prepositions}.
The patterns we see in these prompts suggest that the early representations of the text encoder behave like a ``bag of concepts'', with a representation for each concept but no clear relations between them.

\begin{table}
\resizebox{\linewidth}{!}{%
 \centering
 \begin{tabular}{l c c c c}
 \toprule
 
 & \multicolumn{2}{c}{Antecedent first} & \multicolumn{2}{c}{Antecedent second} \\
 \cmidrule(lr){2-3}\cmidrule(lr){4-5}
 Model & $1^{\text{st}}$ noun & $2^{\text{nd}}$ noun & $1^{\text{st}}$ noun & $2^{\text{nd}}$ noun \\
 \midrule
 DF (T5) & 50.8\%& 33.87\% & 35.50\% & 51.60\% \\
 SD (Clip) & 58.4\%& 23.80\% & 54.90\% & 27.90\% \\
 \bottomrule
 \end{tabular}
 }
\caption{The percentage of prompts in each group where the antecedent noun (either the first or the second noun mentioned) appeared earlier.}
\label{tab:precedence}
\end{table}

\begin{figure}[th]
    \centering
    \includegraphics[width=\linewidth]{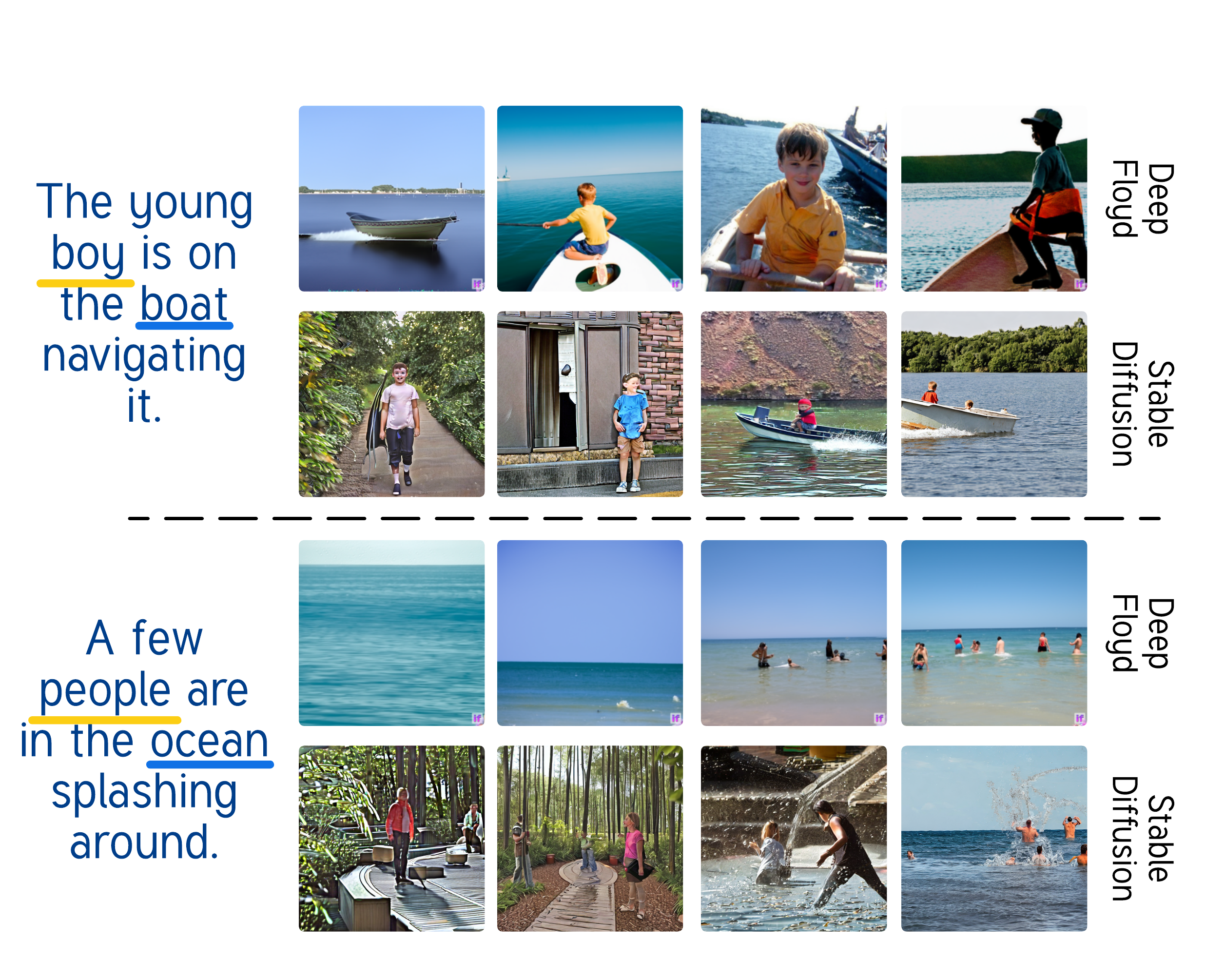}
    \caption{Difference between DF and SD models. The antecedent is marked in yellow and the descendant is marked with blue. In many cases, the antecedent appears in an earlier layer in DF, while the first noun tends to appear first in SD, regardless of its syntactic role.}
    \label{fig:syntactic_dependencies}
\end{figure}

\subsection{Syntactic dependencies}
\label{sec:syntax}

To investigate the order in which different objects emerge, we focus on the association between syntactic depth and the appearance order of nouns. Specifically, we explore whether, in a dependency path where noun A precedes noun B, noun A appears at earlier layers through \method.
Using 63K prompts from COCO that we parsed with Stanza \cite{qi2020stanza}, we filtered for instances with two nouns per prompt and analyzed the dependency relations between the nouns. We categorized the data based on the linear position of the antecedent and generated images with 40 random samples from each group. For each generation and intermediate layer, and each object X, we queried whether object X appears in the image.

\paragraph{Results.}

First, we sometimes observe a ``race'' between the nouns: in 11.9\% of the cases in DF, the object that appears in an earlier layer disappears at a later layer, while the other object takes dominance. See Appendix \ref{app:race} for examples.

Second, 
Table~\ref{tab:precedence} presents information on the order of generation for both models, revealing that
the sequence in which objects emerge during the computation process is determined by either their linear or their syntactic precedence, \emph{depending on the particular text encoder}.
In DF's T5 text encoder, slightly over half of the instances feature the antecedent appearing at an earlier layer than the descendant, with a smaller fraction showing the opposite, and the rest indicating simultaneous appearances.
This holds true regardless of linear order. Conversely, in SD's CLIP, the first noun tends to appear before the second more frequently, irrespective of the syntactic role. See Figure \ref{fig:syntactic_dependencies} for a qualitative example of this case. %

While the two models differ in multiple respects (architecture, pretraining data, training objective, and more), 
it is intriguing to observe that T5, trained on a language modeling objective, demonstrates a greater awareness of syntactic structure compared to CLIP -- a model trained to align pairs of prompts and images without a specific language modeling objective.
This discrepancy points to a possible impact of training objectives on the models' representation building process.

\section{Memory Retrieval}
\label{sec:memory}

Text-to-image diffusion models are able to retrieve information of many concepts \cite{ramesh2022hierarchical}, encompassing entities like notable individuals, animals, and more.
Memory retrieval---the recall of stored information---involves a constructive process rooted in the interactive dynamics between memory trace features and retrieval cue characteristics \cite{smelser2001international}.
In this section, we leverage the \method~to scrutinize the memory retrieval mechanism in the text encoder.

\begin{figure}
 \centering
 \includegraphics[width=\linewidth]{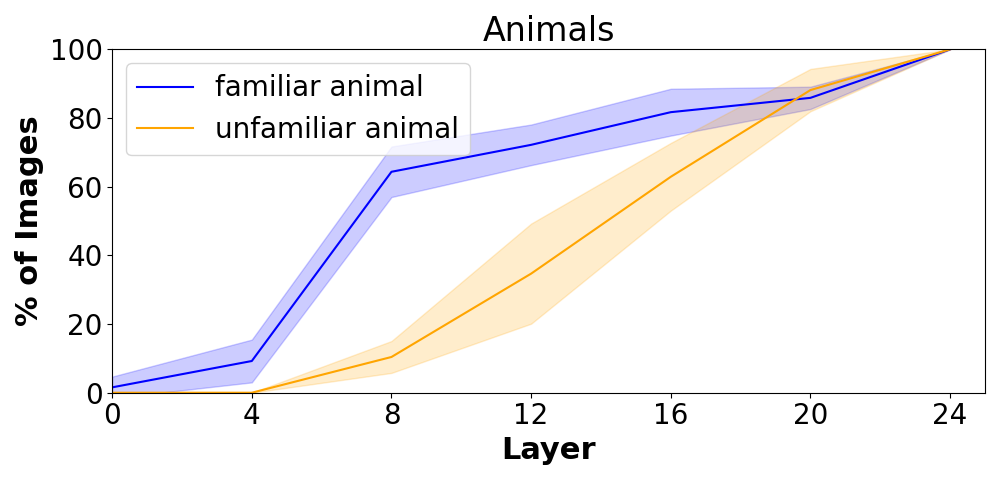}
 \caption{Common vs. uncommon animals across layers. Common animals emerge at much earlier layers.}
 \label{fig:common_vs_uncommon_animals}
\end{figure}

\begin{figure}[t]
\centering
 \includegraphics[width=\linewidth]{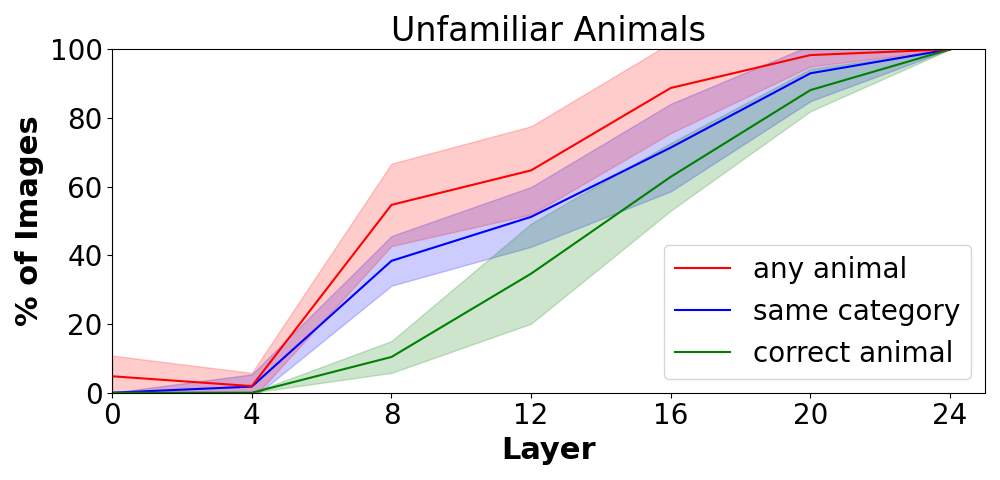}
 \caption{Subset of layers encoding different features in the process of uncommon animal generation.}
 \label{fig:common_uncommon_df}
\end{figure}

\subsection{Common and Uncommon Concepts}
We investigate whether there is a difference in the generation process for prompts describing common and uncommon concepts, using 
a list of common and uncommon \emph{animals}.\footnote{We use animal species as we found that models can correctly generate images of uncommon species, unlike uncommon objects and celebrities.} 
Commonality in this context does not refer to the commonality of an animal in the world, but rather to its commonality in the training data.
As a proxy to measure commonality in the training data, we utilized the average daily view statistics of Wikipedia pages %
from October 2022 to October 2023. 
An animal was deemed ``common'' if it had an average of 1500 visits per day on its Wikipedia page (e.g., kangaroo), while one having fewer than 800 visits per day was deemed ``uncommon''.
We  verified this distinction by examining the frequencies of species names in the LAION2B-en dataset \cite{schuhmann2022laion}, extracted by \citet{Samuel2023SeedSelect}, and found that the frequency of common species was greater than that of uncommon species with statistical significance (Appendix \ref{app:animal_experiment}).

Since the model may have seen the uncommon animals less frequently during training, their generation may take longer. We annotate each image by asking if the specific animal in the prompt appears in the generated image.

\paragraph{Results.} As summarized in Figure \ref{fig:common_vs_uncommon_animals},
\emph{common concepts emerge early}, as early as layer 8 out of 24. In contrast, \emph{uncommon concepts gradually become apparent across the layers}, with  accurate images generated primarily at the top layers.

\begin{figure}[t]
\centering
 \includegraphics[width=\linewidth]{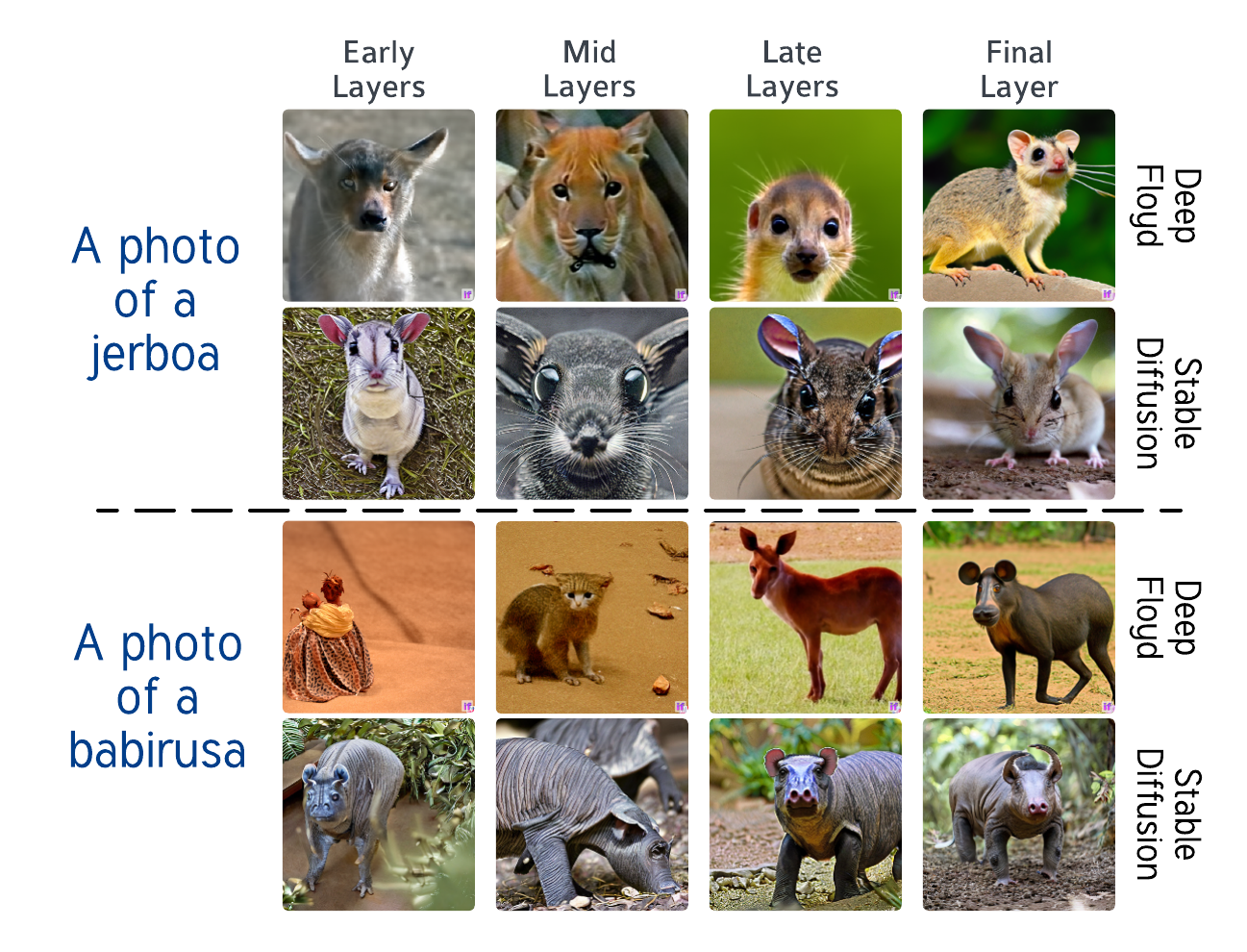}
 \caption{Incremental progression in DF versus early knowledge representation in SD.}
 \label{fig:gradual_retrival_sd_df}
\end{figure}

\subsection{Gradual Retrieval of Knowledge}

To delve deeper into the knowledge retrieval process, we pose additional questions for uncommon animal:
(a) Is there an animal in the image?
(b) Does the image feature an X? where X is the informal ``category''\footnote{We chose to use an informal taxonomy as the animal kingdom taxonomy is a complex subject under research and debate, and its terms are not common to the general population and, hence, likely less present in the T2I training data.} of the animal, such as ``mammal'' and ``bird''.
(c) Does the image depict the exact animal in the prompt?

\paragraph{Results.}

Figure \ref{fig:common_uncommon_df} illustrates \emph{incremental knowledge extraction}, beginning with a general animal, progressing to a more specific animal within the same category, and reaching a representation of the particular animal mentioned in the prompt.

Though the plot for SD reveals a similar pattern (Appendix \ref{app:sd_results}), qualitative analysis reveals \emph{distinct knowledge retrieval patterns between the two models}:
In the case of DF's T5, knowledge retrieval is gradual, unfolding as computation progresses (Figure \ref{fig:gradual_retrival_sd_df}). Layers generate animal, mammal, and ultimately construct a representation of the specific animal.
However, SD's text encoder, Clip, does not display a similar progression of retrieval. The model establishes the representation less gradually: The first layer with a meaningful image already closely resembles the final animal, with subsequent layers mainly refining its characteristics.
These differences echo the syntactic findings in Section \ref{sec:syntax}. They suggest that pretraining objectives, data, or model architecture might influence information organization, leading to distinct memory retrieval patterns.

\vspace{3pt}
\subsection{Gradual refinement of features}

\begin{figure}
 \centering
 \includegraphics[width=\linewidth]{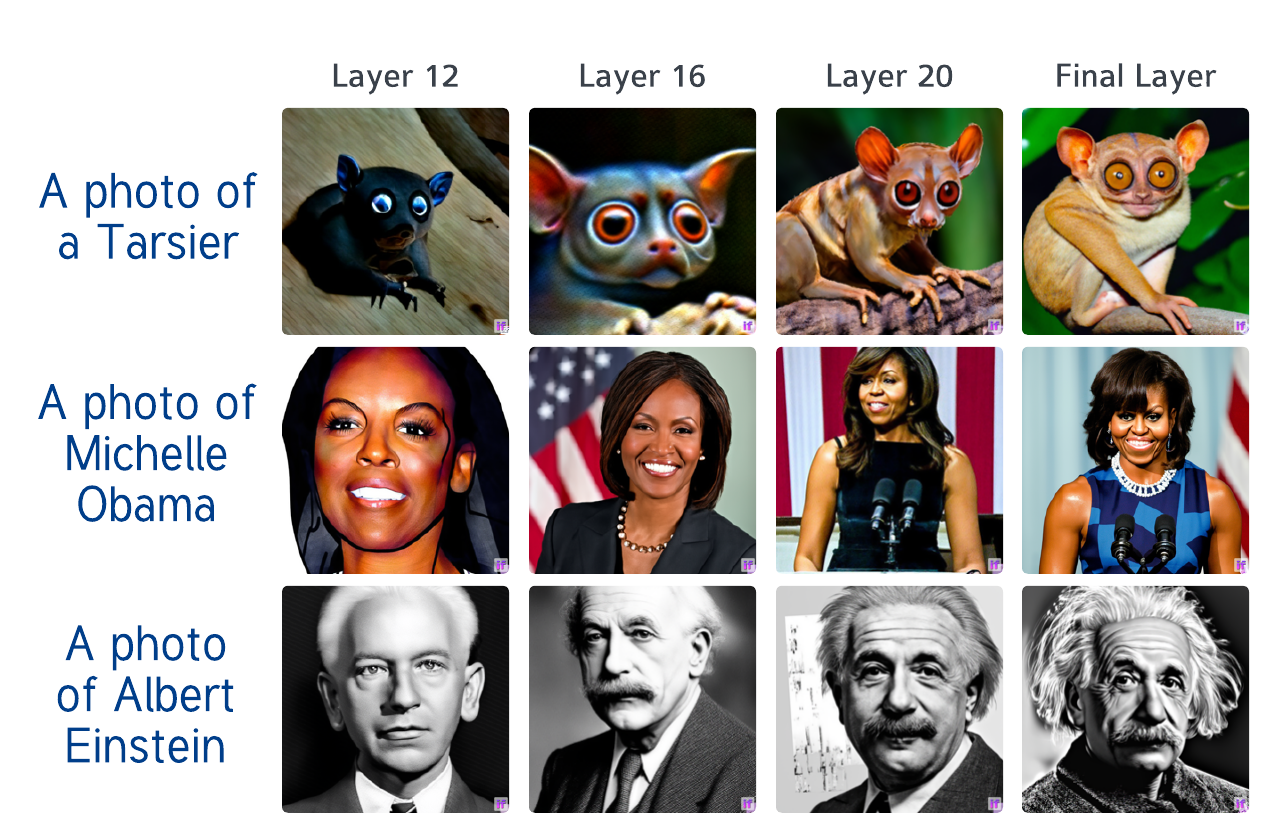}
 \caption{Intricate details are refined gradually.}
 \label{fig:gradual_refinement}
\end{figure}

As the computation progresses, both accuracy and realistic representation significantly improve with refining details at each step. This progression is evident in Figure \ref{fig:gradual_refinement} (top row), as seen in the gradual refinement of the ``Tarsier'' image. A similar trend occurs in the representation construction of human subjects, with facial features undergoing refinement for a more faithful portrayal (Figure \ref{fig:gradual_refinement}, rows 2+3). To systematically assess this phenomenon, we compiled a list of 30 celebrities, using \method~to generate images from intermediate representations in the text encoder.
For each prompt and generated image, we ask:
(a) Is there a person in the image?
(b) Does the person align with the celebrity's (self-identified) gender? %
(c) Does the person exhibit the celebrity's style (hair, clothing, etc.)?
(d) Is the individual in the image distinctly recognizable as the specified celebrity based on facial features?

\begin{figure}
 \centering \includegraphics[width=\linewidth]{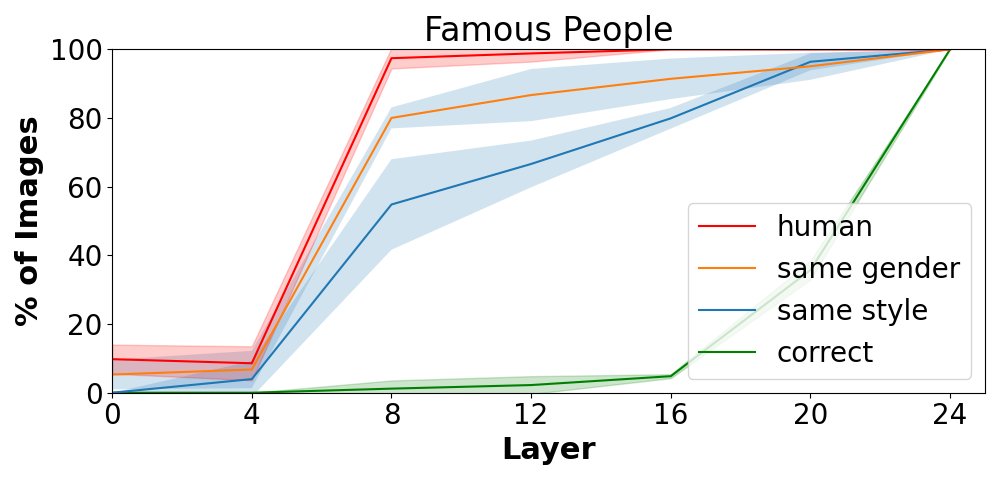}
 \caption{The distribution of feature granularity across layers in generated images.}
 \label{fig:famous-plot}
\end{figure}

\vspace{-3pt}
\paragraph{Results.} Figure \ref{fig:famous-plot} quantifies the \emph{step-by-step construction of the representation}, culminating in its maximum resemblance to the celebrity. The integration of distinct features follows a hierarchical pattern, progressing from broad characteristics (such as the overall human form) to finer details (specifically, facial features), which become evident only in the final layers.

\paragraph{Discussion.}
Our results on the gradual retrieval and refinement of knowledge suggest an alternative perspective on how knowledge is encoded in language models.
This viewpoint is different from recent work suggesting that models utilize a key--value memory structure, where facts are local to specific layers \cite{geva-etal-2022-transformer, meng2022locating, arad2023refact}.
Our results indicate that some information is distributed across layers, allowing for a gradual retrieval of knowledge rather than a retrieval at a particular point in the model. This aligns with earlier research proposing hierarchical representations in vision models \cite{10.1007/978-3-319-10590-1_53, 
zhou2014learning, bau2017network}.

\section{Analyzing Model Failures}
\label{sec:error_analysis}
\vspace{-3pt}

\begin{figure}
 \centering
 \includegraphics[width=\linewidth]{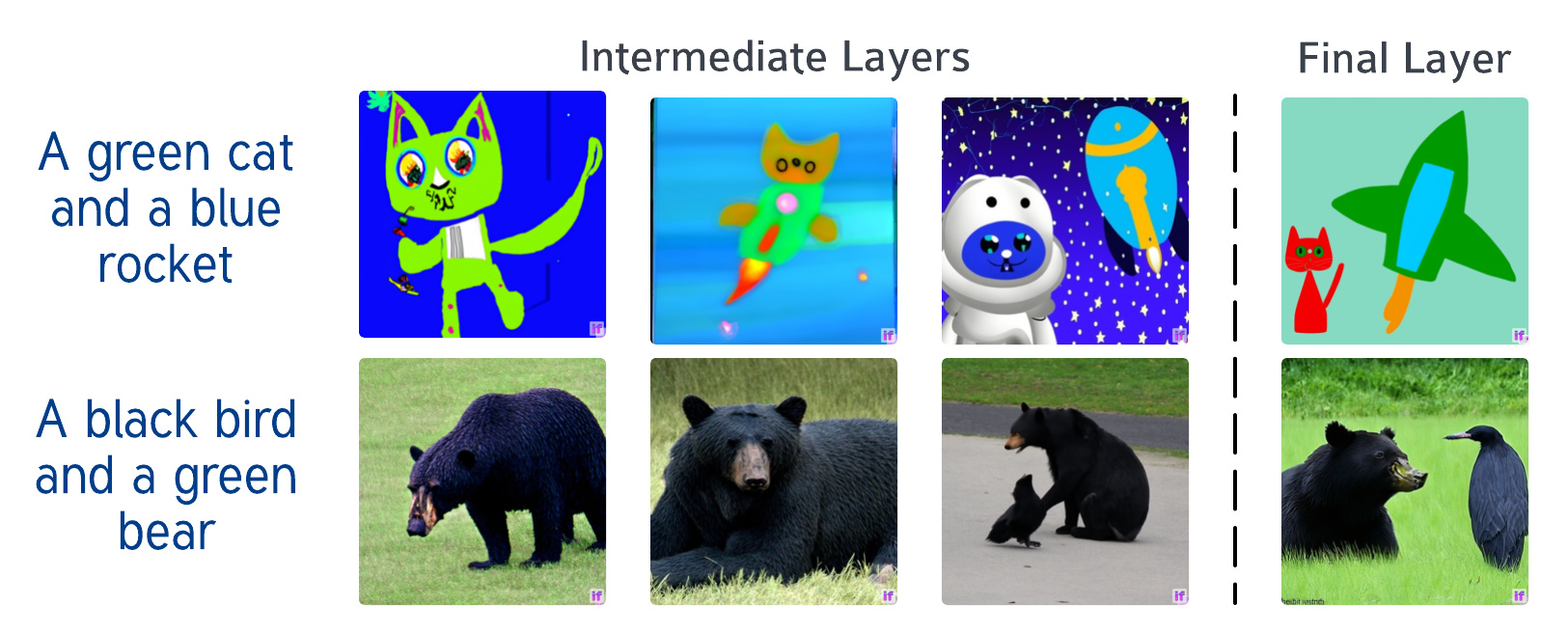}
 \caption{Examples of failure cases of T2I models (right). Using the \method~(left) we can observe different patterns. In the first case (top row), the model is able to correctly generate each entity separately, but fails to combine them in the final layer. On the other hand (bottom row), the model is unable to generate a green bear in any of the intermediate representations.}
 \label{fig:fail_example}
\end{figure}

\begin{figure}[t!]
 \centering
 \includegraphics[width=1\linewidth]{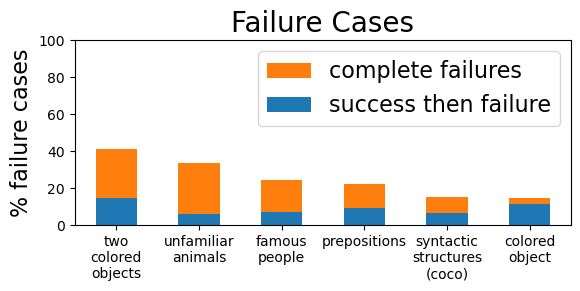}
 \caption{ %
 Many cases display successful generations from earlier layers before turning into failures. }
 \vspace{-1em}
 \label{fig:error_analysis}
\end{figure}

In this section, we delve into cases where the T2I model fails to generate images that align with the input prompt.
First, we investigate all failure cases in all experiments. Figure \ref{fig:error_analysis} shows the percentage of failures for each experiment that had over 10 failures. We split failures to two types: \emph{complete failures} where no layer generated a correct image through \method{}, and cases where at least one layer generated a correct image, but the top layer led to a failure (\emph{success then failure}). 

\begin{figure}[t!]
 \includegraphics[width=\linewidth]{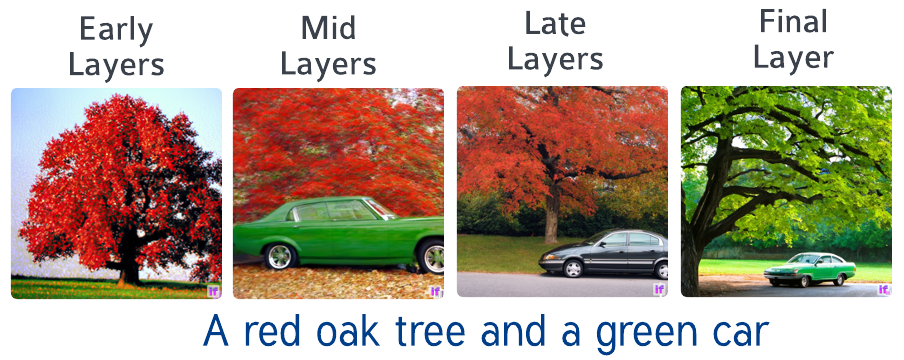}
 \vspace{-1.5em}
 \caption{\method~reveals a correct image generation at a middle layer, while the final image fails to fully represent the prompt.}
 \vspace{-1.2em}
 \label{fig:error_analysis_qualitative}
\end{figure}

Generally, the percentage of failure cases (total height of each bar) is low, from 10\% to 25\% for most categories. Prompts about two colored objects have a higher failure rate. 
Importantly, in many failure cases, the representations in earlier layers lead to a correct generation via our method. Notably, in simple prompts (relations and colored objects), about 80\% of the failures had successful generations at earlier layers -- see Figure~\ref{fig:error_analysis_qualitative} for an example. 
Once more constraints are imposed (two colored objects), we have a lower rate of early success. 
Finally, for knowledge-related tasks (famous people, unfamiliar animals), there are very few cases of early success turned to failure. 
Presumably, when the model fails, it is mostly because it does not encode the information at all.

Next, we zoom in on prompts describing two entities with different colors, as these prompts lead to the highest failure rate in our experiments. 
Examples for failure cases are shown in figure \ref{fig:fail_example}.
Examining the final layer output images, the failures look similar: in both cases one or more entity was generated in the wrong color.
However, using the \method, we reveal two different failure patterns:
In the first example (``A green cat and a blue rocket''), both the blue rocket and the green cat are generated separately successfully in intermediate layers, while final output image fails to combine them into a single image. 
This suggests that the failure stems from an unsuccessful combination of the two concepts.
In the second example (``A black bird and a green bear''), the bear consistently appears black across all intermediate layers, signifying that the model struggles to generate a green bear throughout the encoding process of the text.
A possible explanation is that black bear is a type of animal, which might mean that the phrase ``black bear'' is common in the training data, thus the appearance of the phrase ``black'' in the prompt biases the model towards generating black bears.
This analysis reveals two different sources of failures that occur in compound prompts: either (1) the model fails at coupling a particular concept and color because it is biased towards another color, or that (2) the model can successfully couple each concept and color but fails to combine them.

We analyze how frequently each source of failure occurs, focusing on prompts that failed to generate a correct image from the final layer in at least 75\% of cases.
For each entity, we count the number of times it appeared in the correct color in at least one early layer.
We find that for 40\% of the failure cases in DF (70\% in SD), at least one of the entities did not appear at all in images from earlier layers (type 1). The remaining set of failures had the correct color for each of the objects appear at some point in the computation (type 2).

\section{Related Work}
\label{sec:related_work}
\vspace{-4pt}

\paragraph{Interpreting language models.}
A wide range of work has analyzed language model internals. We briefly mention a few directions and refer to existing surveys \cite{belinkov2019analysis,Rogers2020API,madsen2022post,ferrando2024primer}. 
Probing classifier are used to analyze whether internal representations correlate with external properties \cite[e.g.,][]{ettinger-etal-2016-probing,hupkes2018visualisation}. However, probing has various inherent flaws such as memorization \cite{belinkov2022probing},
and requires costly annotations for fine-grained analysis like the visual characteristics of a specific animal species or person.
Interventions in representations measure how they impact a model's prediction \cite[e.g.,][]{vig:2020:neurips,elazar2021amnesic,meng2022locating}, and while they offer powerful insights, they are also challenging to design \cite{zhang2023towards} and limited in scope.
In contrast, the \method~ proposes a simple yet generic mechanism to visually interpret intermediate representations without requiring additional data or training, enabling exploration of fine-grained visual features.

Another influential approach is the Logit Lens \cite{nostalgebraist}, which projects intermediate representations of language models onto a probability distribution over the vocabulary space. The logit lens captures the internal computation of the language model, and the flow of information across modules \cite{geva-etal-2022-transformer, katz2023interpreting, pal-etal-2023-future}. 
This line of work has focused on auto-regressive decoder language models.
Inspired by this idea, we propose using the diffusion module in T2I pipelines to visualize intermediate prompt representations, revealing the text encoder's computation process.

\vspace{-3pt}
\paragraph{Interpreting vision-language models.}

Compared to unimodal models, research on interpretability in multimodal vision-language models is rather limited. 
\citet{goh2021multimodal} found multi-modal neurons responding to specific concepts in CLIP \cite{clip2021} and \citet{gandelsman2023interpreting} decomposed CLIP's image representations into text-based characteristics.

\citet{tang2022daam} were the first to propose a method to interpret T2I pipelines, by analyzing the influence of input words on generated images via cross-attention layers. 
\citet{chefer2023hidden} decomposed textual concepts, focusing on the diffusion component. 
In contrast, our work investigates the under-explored text encoder in T2I pipelines. 
Unlike previous methods, the \method~reveals gradual processes within the model, not focusing only on the final output.

\section{Discussion and Conclusion}
\label{sec:discussion}

We introduce the \method, a novel method to analyze language models within T2I pipelines.
Our approach deconstructs the T2I pipeline by examining the output of each block within the text encoder, thereby providing a deeper insight into language-to-visual concept translation.
We are the first, to our knowledge, to propose a method to interpret the text encoder and its internal computation process in the context of T2I models.
Given that the text encoder is a crucial component of T2I models, enhancing its interpretability contributes to a deeper understanding of the entire generation process.
We showcased the method's potential by analyzing two open-source text encoders used in T2I pipeline across diverse topics.

Our work contributes to a growing body of work on analyzing how models process information across various components.
The \method\  may have many potential applications as a first method of its kind, including similar applications to prior interpretability techniques such as improving model efficiency \cite{din2023jump, dalvi-etal-2020-analyzing2} and tracing factual associations in language models, facilitating more accurate model editing methods \cite{meng2022locating, arad2023refact}.

Other future directions using the \method~may aid in identifying points of failure in the computation process or remove undesired traits from early layers such as hallucinations, toxicity, or incorrect factual knowledge. 
Lastly, while we focused on entire blocks, our approach paves the way for visualizing individual sub-block components such as individual MLPs, attention heads, and residual connections.

\section*{Acknowledgements}
This research was supported by the Israel Science Foundation (grant 448/20), an Azrieli Foundation Early Career Faculty Fellowship, and an AI Alignment grant from Open Philanthropy. HO is supported by the Apple AIML PhD fellowship. DA is supported by the Ariane de Rothschild Women Doctoral Program.

\section*{Limitations}
While the \method~ provides a method to interpret the intermediate representations of the text encoder of T2I models, there are several limitations. 

First, we are limited by the number of publicly available and open source T2I models and their corresponding text encoders. Extending our method to interpret other language models, whether or not they are used in T2I pipelines, offers a promising direction for future research.

Additionally, most of our experiments utilized automatically generated prompts, used to isolate and investigate specific effects. Such synthetic prompts are often less complex compared to prompts written by humans, and follow specific patterns. Although we experimented with a set of natural prompts, further exploration using a wider range of prompts could provide deeper insights into the behavior of text encoders in T2I models.

Lastly, the \method\ requires further annotation in order to derive large-scale conclusions. In this work, we relied on human and automatic annotation to answer questions on specific attributes of the generated images. This limitation stems from using images as the output of our method, however, we believe using images results in richer and more complex interpretations. 

\section*{Ethics Statement}
In this work, our primary objective is to enhance the transparency of text-to-image models. While not the focus our analyses, the \method\ has the potential to unveil biases within these models. We anticipate that our work will contribute positively to the ongoing discourse on ethical practices in text-to-image models. At present, we do not foresee major ethical concerns arising from our methodology.

\bibliography{anthology,custom}
\bibliographystyle{acl_natbib}

\clearpage 

\appendix
\section{Additional Results}

\subsection{Prepositions}
\label{app:prepositions}

\begin{figure}[th!]
    \centering
    \includegraphics[width=\linewidth]{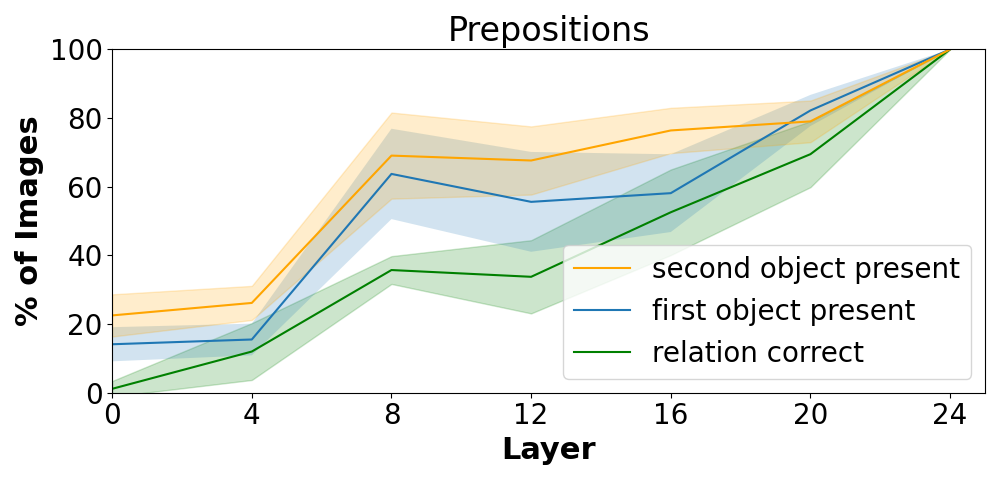}
    \caption{The proportion of images where either the objects, or objects with prepositions, were accurately represented.}
    \label{fig:gradual_complexity_df}
\end{figure}

We explore prepositions in prompts. We investigate how prompts, including certain relations, affect the generation process. These prompts are complex, challenging the compositional understanding of the T2I model. In particular, we examine the prepositions "on" and "in". Figure \ref{fig:gradual_complexity_df} illustrates the percentage of images that correctly generated the concepts for tree categories: each of the objects alone and both objects with the correct relation between them. Our findings reveal that the emergence of each of the objects occurs at an early stage. However, both objects and their correct relation emerge only later in the text encoding.

\begin{figure*}[t]
 \centering
\includegraphics[width=1\textwidth]{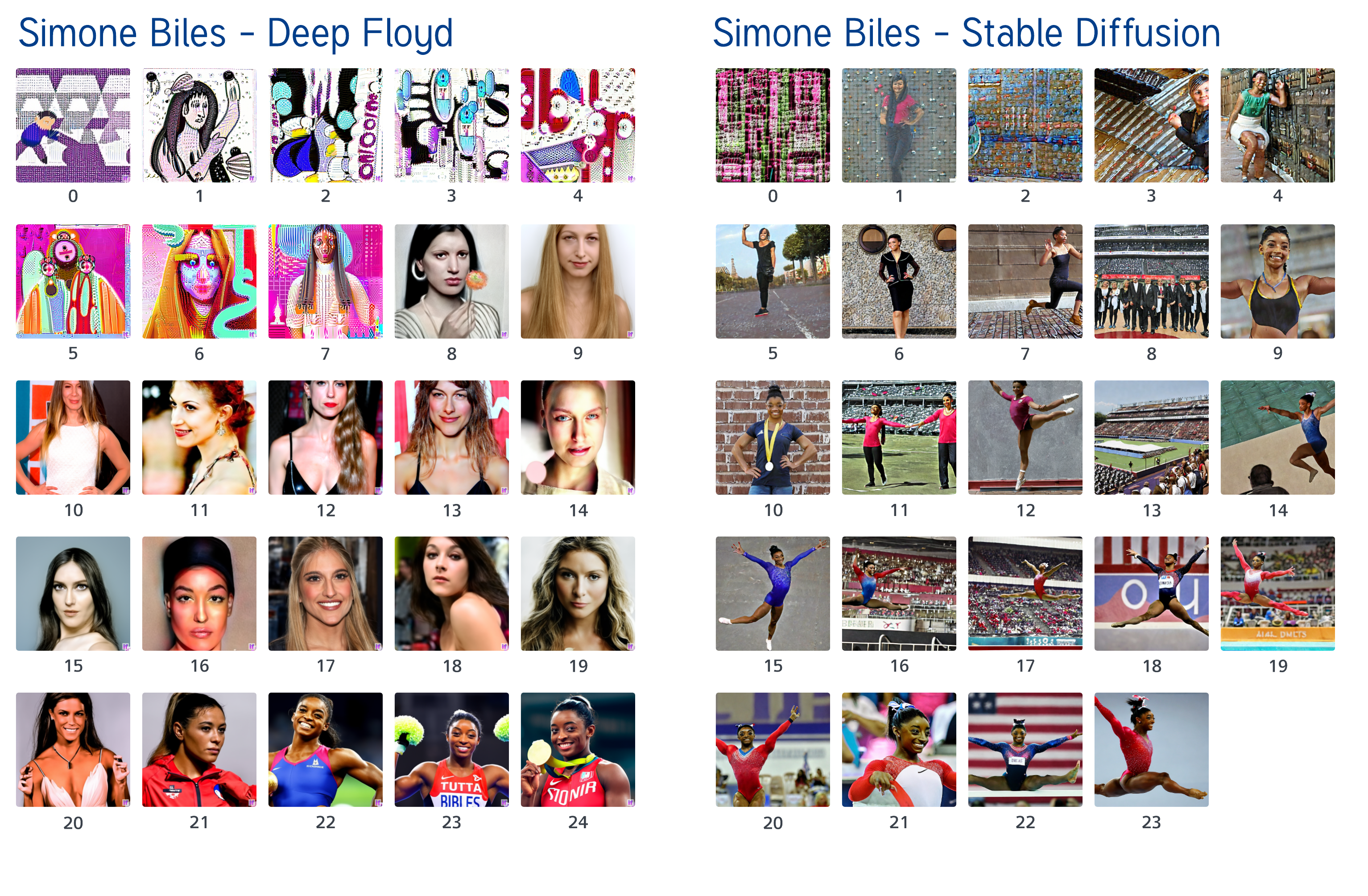}
 \vspace{-1em}
 \caption{Example generations from all layers}
 \label{app:fig:all_layers}
\end{figure*}

\subsection{Race between objects}
\label{app:race}

\begin{figure}[ht]
        \centering
        \includegraphics[width=.75\linewidth]{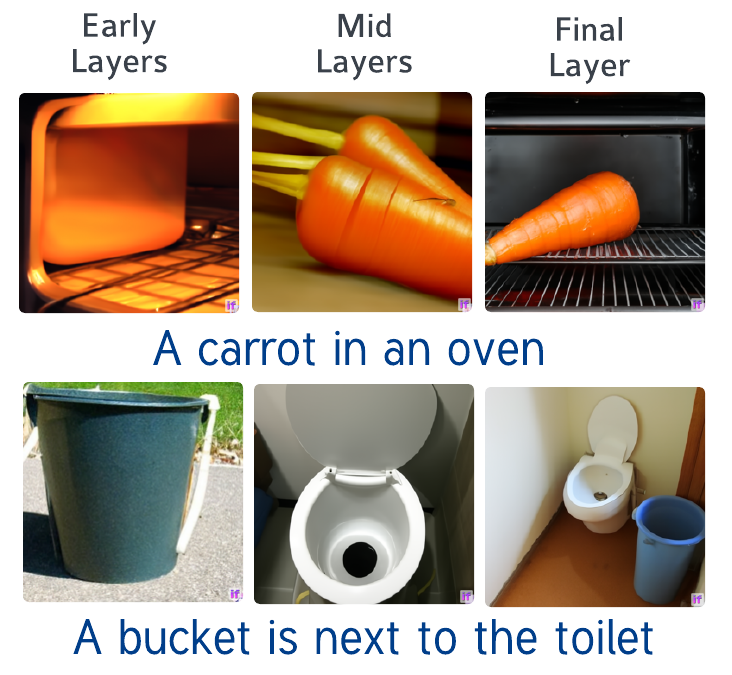}
        \caption{
A sequential ``race'' between two objects in the sentence, where one initially appears before the other, only to subsequently vanish and make room for the latter object.}
        \label{fig:noun_before_noun}
\end{figure}

Figure \ref{fig:noun_before_noun} presents examples of ``race'' between the objects in the prompts: one object appears first, and then disappears at a later layer to make room for the other object, before finally emerging again in the top layers.

\subsection{Final layer norm necessity}
\label{app:final_layer_norm}

\begin{figure}[!htb]
        \centering
        \includegraphics[width=\linewidth]{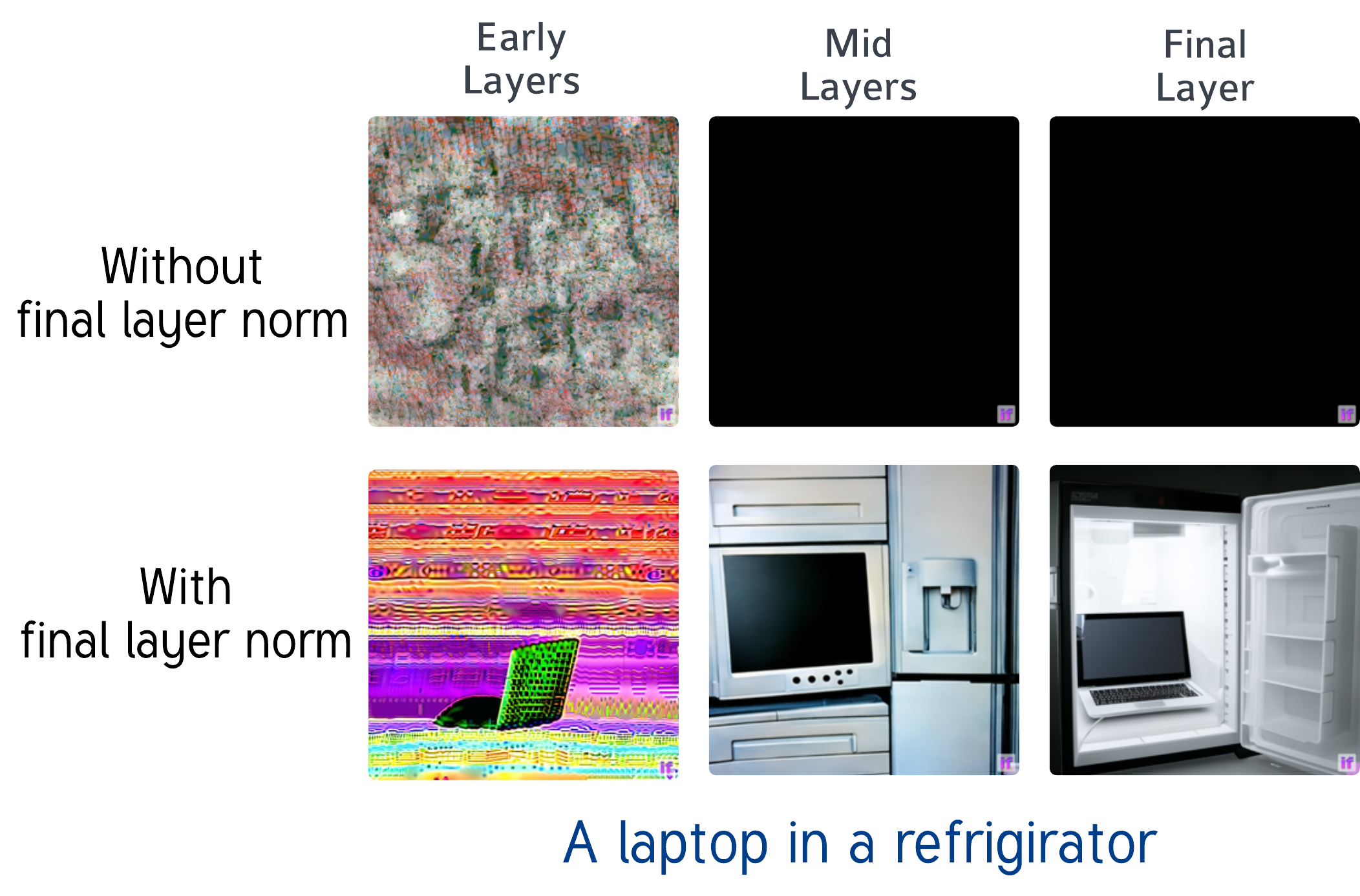}
        \vspace{0.1em}
        \caption{Example generations from \method~with and without the final layer norm. }
        \label{fig:layer_norm_with_without}
\end{figure}

In the \method~process, we pass the output of block $l$ through the last layer norm $ln_f$. However, we examine the option to bypass the $ln_f$ layer and directly connect to the components of the diffusion model. 
As Figure \ref{fig:layer_norm_with_without} demonstrates, images generated without the final layer normalization are meaningless.
The final layer norm thus plays a crucial role in generating meaningful images.
It highlights the necessity of the $ln_f$ layer within \method~pipeline.
A similar finding has been observed in the  LogitLens \cite{nostalgebraist} and TunedLens \cite{belrose2023eliciting}.

\section{Annotation Process}
\label{app:annotation}

The results in this paper rely on human annotators to determine the presence of different concepts in the generated images. We employed a team of ten professional full-time annotators using the Dataloop platform \href{https://console.dataloop.ai/}, in accordance with institutional regulations. The annotator teams was based in India, and were paid a rate of 8 USD per hour, in accordance with laws in India. 

Each annotator received the instructions in Figure~\ref{fig:guidelines}. 
\begin{figure}[t]
\fbox{
\footnotesize 
\parbox{0.46\textwidth}{
On this project, you will have to annotate sets of 50 images. For each set, you will have a yes or no question.
The questions are written at the start of each task name. They end with a “?”. The latter part of the name is in “[ ]” and is not relevant for the questions.
For convenience, we start the question with the statement itself, therefore “dog in the image?” means “Is there a dog in the image?”
The questions vary from simple questions like “Is there a dog in the image?” to more complicated questions like “Is there a red bird on a green boat?”. The images are generated by AI, and might not be realistic. You should answer if the image might be interpreted as the question asks. Examples at the end of this file. 
}}
\caption{Annotation guidelines.}
\label{fig:guidelines}
\end{figure}
The annotators were given the instruction to be liberal towards a positive answer. We manually validated each question, making sure the concepts in the question are not abstract (e.g., ``beautiful''), and that the answer should be clear for each case.
For each experiment, we duplicate 10\% of the images, and ask an additional annotator the same questions, used to calculate inter annotator agreement. For experiments containing rare animals and celebrities, annotators were given reference images from google.

 We provide our main results based on the human annotations. We chose to use human annotations since the existing automatic methods are limited. CLIP as an image classifier was shown to fail when required to explicitly bind attributes to objects \cite{ramesh2022hierarchical, yamada2022lemons}, and exploratory experiments we performed with BLIP \cite{li2023blip} showed similar issues.

  We found a high agreement between GPT-4V \cite{openai2023gpt4} and the human annotators on most tasks and questions, as shown in Table \ref{tab:annotations_agreement}.
  For one experiment -- two colored objects -- we found a high variance using the human annotations and thus extended it to further annotations using GPT4-V.

\begin{table*}[t]
 \centering
\resizebox{\linewidth}{!}{
    \centering
    \begin{tabular}{l c c c c c c}
    \toprule
    
 & \multicolumn{3}{c}{Inter annotator agreements} & \multicolumn{3}{c}{Agreements with automatic annotations} \\
    \cmidrule(lr){2-4} \cmidrule(lr){5-7}
    Question type & \#annotations & f1 & cohen's kappa & \#annotations & f1 & cohen's kappa \\
     \midrule
     One object presence (out of 2) & 416 & 72.5\% & 48.2\% & 1381 & 80.6\%  & 63.8\% \\
     Relation correct & 208 & 73.7\% & 61.4\% & 1319 & 81.3\%  & 70.1\%  \\
     One Color presence & 208 & 76.9\% & 60.7\% & 1671 & 85.3\%  & 85.9\%  \\
     Familiar animals presence & 52 & 94.7\% & 87.2\% & 789 & 85.5\%  & 67.2\%  \\
     Unfamiliar animals presence & 104 & 84.6\% & 81.3\% & 1019 & 84.3\% & 72.4\%  \\
     Unfamiliar animals class presence & 260 & 73.2\% & 59.5\% & 1012 & 91.2\%  & 81.3\% \\
     Syntactic structures correct (coco) & 357 & 80.6\% & 69.7\% & 2962 & 80.0\%  & 59.5\%  \\
     \bottomrule
    \end{tabular}
    }
\caption{A table of agreement between human annotators (left) and between human and automatic annotations averaged over both models. Overall, we see a high agreement between the human annotators and between the human and automatic annotations. For human agreement - the lowest Kappa score is for one object presence, probably due to the ambiguity in early layers, where there is a mix of both objects. For example in fig \ref{fig:gradual_complexity}, second line, layer 12.}
\label{tab:annotations_agreement}
\end{table*}

\section{Animals Experiment: Implementation Details}
\label{app:animal_experiment}

\subsection{Animal classes used}

To measure the gradual knowledge retrieval, one of the questions we ask in the experiment on unfamiliar animals is whether the image contains an animal of class X, where we vary X according to an informal, popular taxonomy that the specific animal belongs to.
Note that although it does not faithfully represent the scientific view on the animals we generate, it is more suitable to observe a model that was trained on data that was taken from the wide internet.

To verify the distinction between familiar and unfamiliar animal species we preformed a Mann-Whitney U rank test \cite{mann1947test} on the frequencies of species names in the LAION2B-en dataset \cite{schuhmann2022laion}, commonly used in the training process of T2I models which was computed by \cite{Samuel2023SeedSelect}. We found that the frequency of familiar species was greater than that of unfamiliar species with a confidence level of 95\%.

\subsection{The full list of animals}

Familiar animals: Beagle, German Shepherd, Labrador Retriever, Dachshund, Bulldog, Ragdoll, Kangaroo, Chicken, Owl, Eagle, Salmon, Catfish, Cod, Orca, Komodo dragon, King cobra, Platypus, Narwhal, Ostrich, cougar.

Unfamiliar animals: Aye-aye,  Dik-dik, Tarsier, Gerenuk, Jerboa, Babirusa, Saola, Galago, Vervet, guppy, Celestial Pearl Danio, Herring, Pike, Walleye, Grebe, Spoonbill, Bee-eater, Taipan, ,Copperhead, Anilius, Skink, Bearded Dragon, Ladybug, Scarab, Blue morpho, Cloudless sulphur, Giant anteater

\section{Implementation Details}
\label{app:implementation}
We implemented our code using Pytorch \citep{pytorch} and Huggingface libraries \citep{wolf2020transformers, von-platen-etal-2022-diffusers}. For each experiment, we generated four images (different seeds) for each layer, and we report the standard deviation over the seeds in all plots.
We use Stable Diffusion v2-1 (CreativeML Open RAIL++-M License) \citep{rombach2022high} and Deep Floyd (DeepFloyd-IF-License) \cite{deepfloyd}. We ran the experiments on the following GPUs: Nvidia A40, RTX 6000 Ada Generation, RTX A4000 and GeForce RTX 2080 Ti.

Our code is available in the supplementary material.

\subsection{Dependency parsing implementation}
\label{app:dpendency_implementation}
We conducted a syntactic structure analysis using Stanza \cite{qi2020stanza}, a Python package. Stanza provides tools for obtaining parts of speech (POS) and syntactic structure dependency parse. To perform this analysis, we executed a Stanza pipeline designed for English. This pipeline returns the tokenized form, POS, lemmatization, and syntactic dependency parsing for a given prompt. We didn't customize any additional parameters and utilized the default settings during the analysis.

\section{Results on Stable Diffusion}
\label{app:sd_results}

\begin{figure}[th!]
    \centering
    \includegraphics[width=\linewidth]{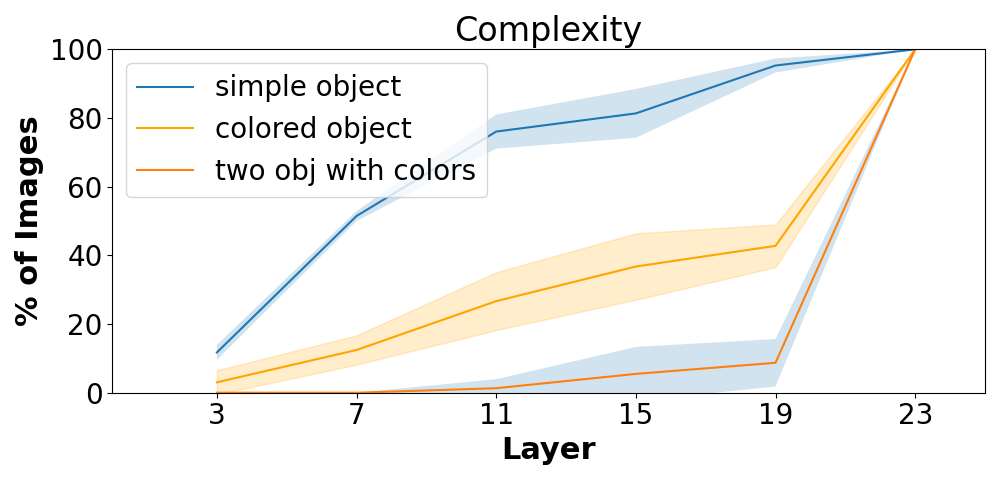}
    \caption{
   [Stable Diffusion] The percentage of images, from each category, for which the prompt matches the generated image, across different intermediate layers. 
    }
    \label{fig:complex_comes_later_sd}
\end{figure}

\begin{figure}[th]
    \centering
    \includegraphics[width=\linewidth]{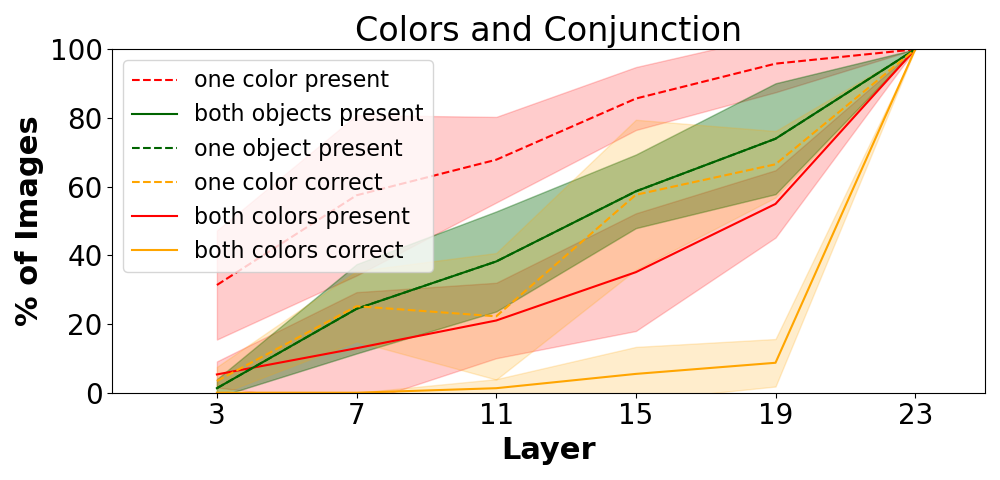}
    \caption{[Stable Diffusion] The proportion of images where either the object, the colors, or both were present, and where either the objects or the colors were accurately represented.}
    \label{fig:gradual_complexity_sd}
\end{figure}

\begin{figure}[t]
    \centering
    \includegraphics[width=\linewidth]{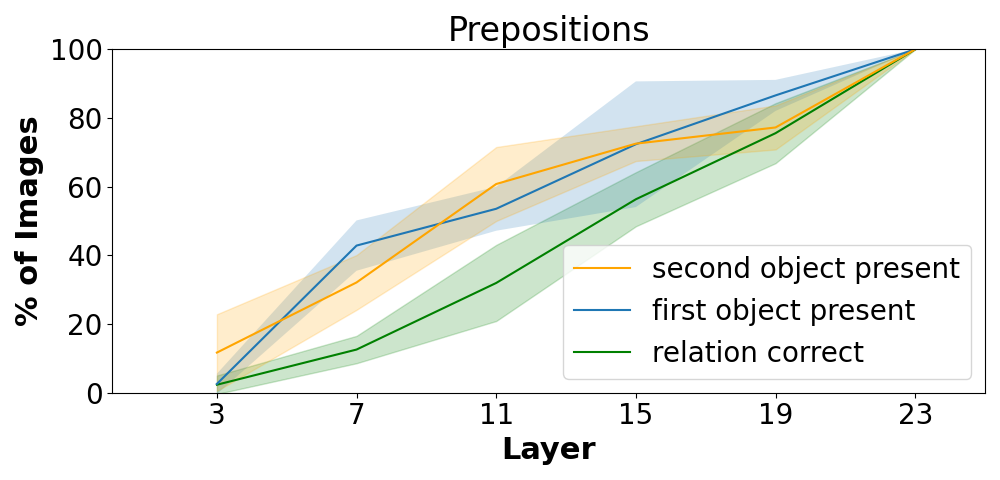}
    \caption{[Stable Diffusion] The proportion of images where either the objects, or objects with prepositions, were accurately represented.}
    \label{fig:prepositions_sd}
\end{figure}

\begin{figure}[t]
    \centering
    \includegraphics[width=\linewidth]{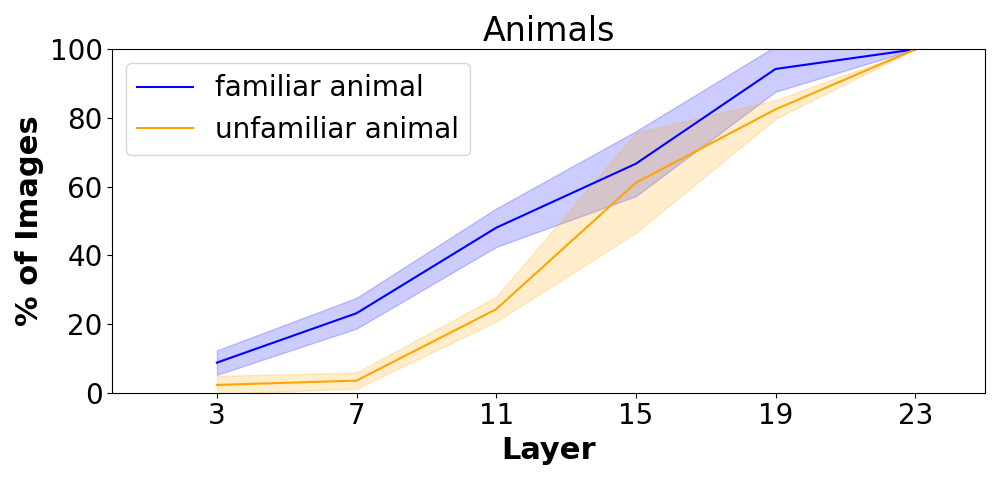}
    \caption{[Stable Diffusion] Familiar vs. unfamiliar animals across layers. 
    }
    \label{fig:common_vs_uncommon_animals_sd}
\end{figure}

\begin{figure}[t]
\centering
    \includegraphics[width=\linewidth]{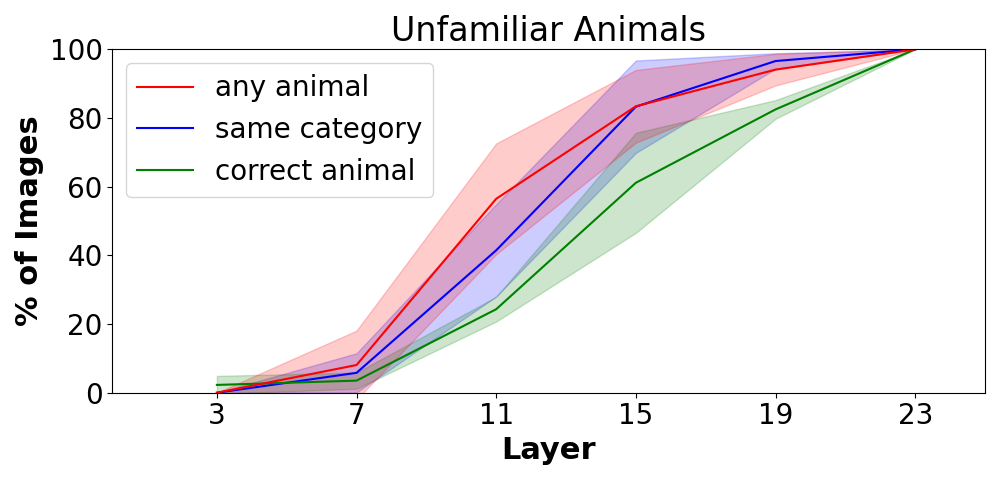}
    \caption{[Stable Diffusion] Subset of layers encoding different features in the process of unfamiliar animal generation.}
    \label{fig:common_uncommon_sd}
\end{figure}

\begin{figure}[t]
    \centering  \includegraphics[width=\linewidth]{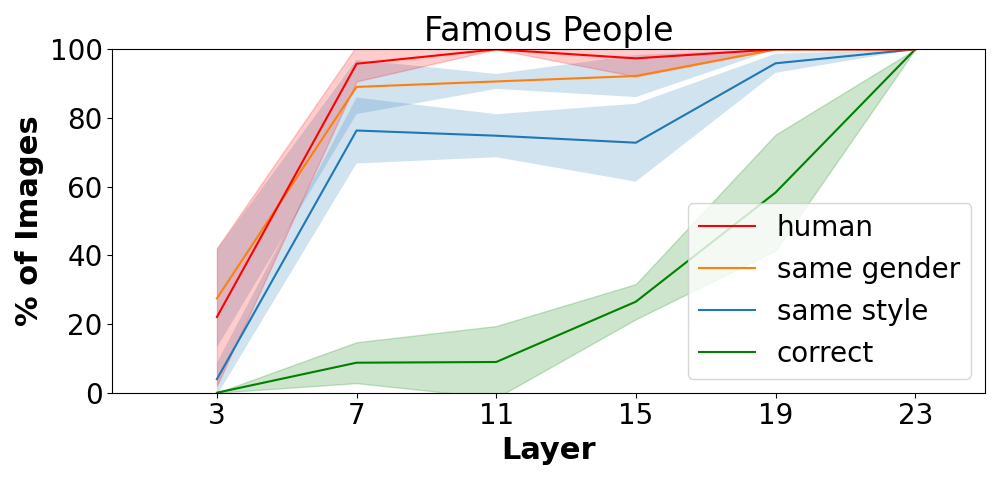}
    \vspace{-1.5em}
    \caption{[Stable Diffusion] The distribution of feature granularity across layers in generated images.}
    \vspace{-1.5em}
    \label{fig:famous-plot-sd}
\end{figure}

\begin{figure}[t]
    \centering
    \includegraphics[width=\linewidth]{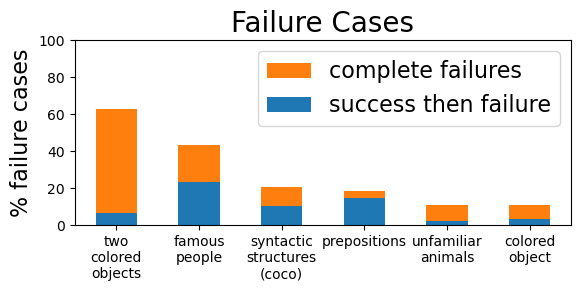}
    \caption{ %
    [Stable Diffusion] Many cases display successful generations from earlier layers before turning into failures. }
    \label{fig:error_analysis_sd}
\end{figure}

To complement the results in the main paper, we provide Figures 
\ref{fig:complex_comes_later_sd}--\ref{fig:error_analysis_sd} from Stable Diffusion. 

\begin{figure*}[t]
 \centering
\includegraphics[width=\textwidth]{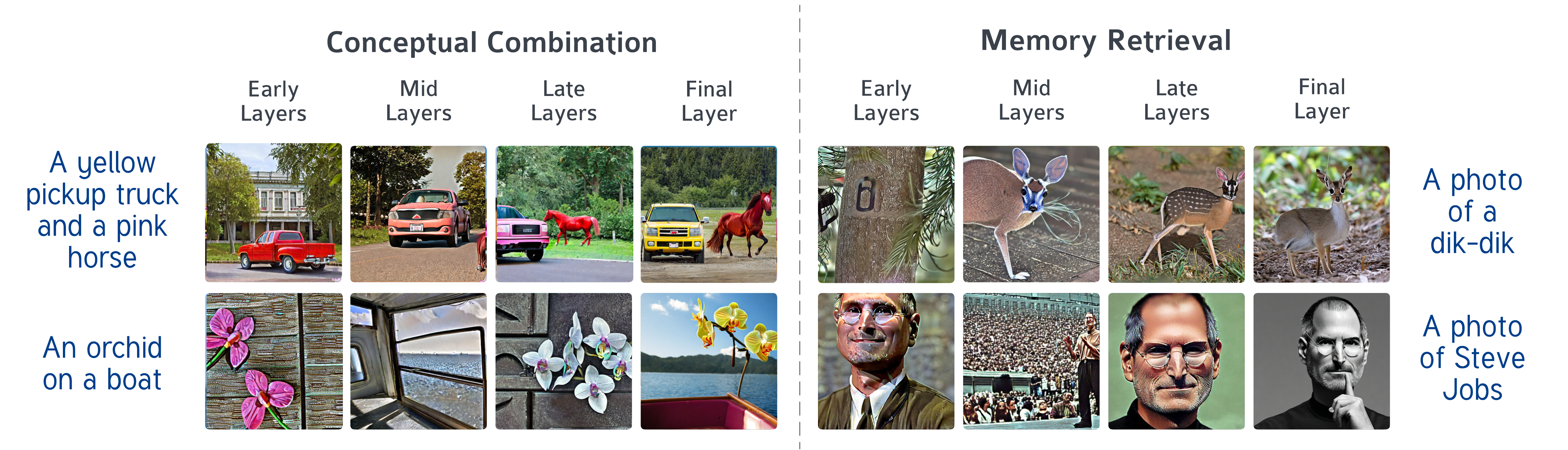}
 \caption{[Stable Diffusion] Insights gained from using \method. \textbf{Conceptual Combination} (left): early layers often act as a ``bag of concepts'', lacking relational information which emerges in later layers. \textbf{Memory Retrieval} (right): concepts emerge early and gradually refine over layers.}
 \label{fig:examples}
\end{figure*}

\end{document}